\documentclass[11pt,a4paper]{article}
\usepackage[hyperref]{emnlp-ijcnlp-2019}
\usepackage{times}
\usepackage{latexsym}

\usepackage{url}
\usepackage{graphicx}
\usepackage{subfigure}
\usepackage{psfrag}
\usepackage{booktabs}
\usepackage{multirow}
\usepackage{mathtools}
\usepackage{placeins}
\usepackage{caption}

\aclfinalcopy 

\newcommand{\defeq}{\vcentcolon=}

\title{X-SQL: reinforce schema representation with context}
\author{Pengcheng He, Yi Mao, Kaushik Chakrabarti, Weizhu Chen \\
  Microsoft Dynamics 365 AI \\
  {\tt \{penhe,maoyi,kaushik,wzchen\}@microsoft.com} }

\date{}
\begin{document}
\maketitle
\begin{abstract}
In this work, we present X-SQL, a new network architecture for the problem of parsing natural language to SQL query. X-SQL proposes to enhance the structural schema representation with the contextual output from BERT-style pre-training model, and together with type information to learn a new schema representation for  down-stream tasks. We evaluated X-SQL on the WikiSQL dataset
and show its new state-of-the-art performance. 
\end{abstract}

\section{Introduction}

Question Answering (QA) is among the most active research areas in natural language processing recently. In this paper, we are interested in QA over structured databases. This is usually done by mapping natural language question to a SQL query representing its meaning, a problem known as semantic parsing, followed by executing the SQL query against databases to obtain the answer.


The largest annotated dataset for this problem is WikiSQL~\citep{zhongSeq2SQL2017}. 
Early work adopts a neural sequence-to-sequence approach with attention and copy mechanism, 
while recent focus has been on incorporating the SQL syntax into neural models. \citet{xu2017sqlnet} and \citet{yu2018naacl} capture the syntax via dependency between different prediction modules. \citet{dong2018acl} and \citet{Finegan-Dollak2018} use a slot filling approach where syntactic correctness is ensured by predefined sketches. 

Recent advances in language representation modeling~\citep{radford2018improving, devlin2018bert} demonstrate the value of transfer learning from large external data source. For WikiSQL, the work of \citet{Hwang2019} has shown significant improvement with such pre-training techniques. 
In view of this trend, we propose X-SQL, an improved pre-training based neural model with contributions from the following three perspectives. 

There are two types of textual information to be considered for this problem: one is the unstructured natural language query, and the other is the structured data schema. Previous work either models them independently or builds cross-attention between them \cite{xu2017sqlnet,Shi2018}. While the structured information such as table column is relatively stable, natural language queries are highly variable. We leverage an existing pre-trained model named MT-DNN \cite{Liu2019} to capture this variation and summarize the unstructured query into a global context representation, which is then being used to enhance the structured schema representation for downstream tasks.  To the best of our knowledge, this is the first attempt to incorporate BERT-style contextual information into a problem-dependent structure, and build a new representation to better characterize the structure information.

Second, part of SQL syntax is bounded to the type of structured data schema. For example, aggregator \textsc{MIN} only appears with numerical column, and operator $>$ doesn't pair with string typed column. We incorporate schema type information in two places. Section~\ref{sec:encoder} describes type embedding with a modification of pretrained language representation, and Section \ref{sec:output} shows how to further improve certain sub-tasks with a separately learned type embedding.

Lastly, we observe previous approach using multiple binary classifiers for where clause prediction cannot effectively model the relationship between columns, since each classifier is optimized independently, and their outputs are not directly comparable.  To tackle this issue, X-SQL takes a list-wise global ranking approach by using the Kullback-Leibler divergence as its objective to bring all columns into a comparable space (Section~\ref{sec:training}). 



\section{Neural Architecture}

The overall architecture consists of three layers: sequence encoder, context enhancing schema encoder and output layer.

\begin{figure*}
\centering
    \subfigure[Context enhancing schema encoder (Equation~\ref{eq:attnpooling}).]{%
        \label{fig:contextreinforcing}
        \includegraphics[width=0.31\textwidth, trim={1mm 1mm 1mm 1mm}, clip]{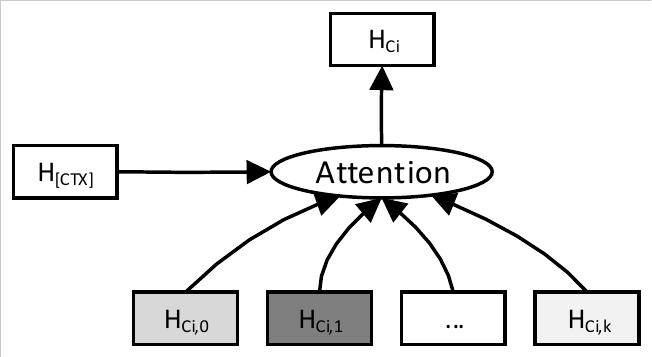}
    }%
    \hfill
    \subfigure[Sub-network that modulates schema representation with context (Equation~\ref{eq:colrep}).]{%
        \label{fig:colrep}
        \includegraphics[width=0.31\textwidth, trim={1mm 1mm 1mm 1mm}, clip]{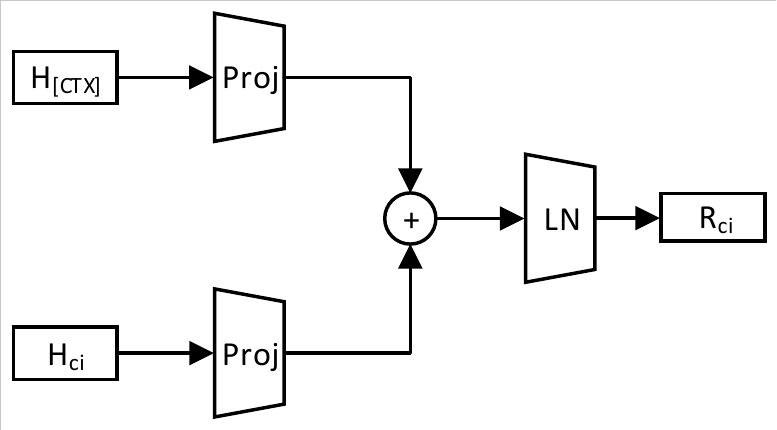}        
    }%
    \hfill
    \subfigure[Neural network for select column aggregator (Task \textsc{S-AGG}, Equation~\ref{eq:s-agg}).]{%
        \label{fig:s-agg}
        \includegraphics[width=0.31\textwidth, trim={1mm 1mm 1mm 1mm}, clip]{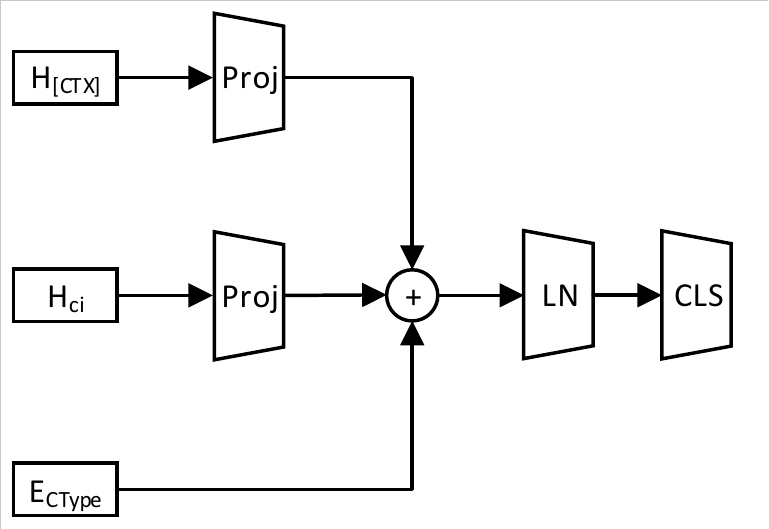}
    }%
\caption{Components of X-SQL neural network model architecture.}
\label{fig:arch}
\end{figure*}

\subsection{Sequence Encoder}\label{sec:encoder}

For the sequence encoder, we use a model similar to BERT~\cite{devlin2018bert} with the following changes:
\begin{itemize}
    \item A special empty column \texttt{[EMPTY]} is appended to every table schema. Its usage will become clear in Section \ref{sec:training}.
    \item Segment embeddings are replaced by type embeddings, where we learn embeddings for four different types: question, categorial column, numerical column and the special empty column.
    \item Instead of initializing with BERT-Large, we initialize our encoder with MT-DNN~\citep{Liu2019}, which has the same architecture as BERT, but trained on multiple GLUE tasks~\citep{Wang2018Glue}. MT-DNN has been shown to be a better representation for down-stream NLP tasks.
\end{itemize}
Note, we rename \texttt{[CLS]} output of BERT to \texttt{[CTX]} in the following sections and Figure~\ref{fig:arch}. This is to emphasize that context information is being captured there, rather than a representation for down-stream tasks.

In addition to these changes, our encoder differs from SQLova with NL2SQL layer~\citep{Hwang2019} in the following important way: while SQLova still runs bi-LSTM/column attention on top of the encoder, our architecture enjoys a much simpler yet powerful design for consequent layers, which we believe is largely attributed to a better alignment of BERT with the problem.

\subsection{Context Enhanced Schema Encoder}\label{sec:context}

Let $h_{\texttt{[CTX]}}, h_{q_1}, \cdots, h_{q_n}, h_{\texttt{[SEP]}}, h_{C_{11}}, \cdots, h_{\texttt{[SEP]}}, \allowbreak h_{C_{21}}, \cdots h_{\texttt{[SEP]}}, \cdots, h_{\texttt{[EMPTY]}}, h_{\texttt{[SEP]}}$ denote the output from the encoder, each of dimension $d$. Each question token is encoded as $h_{q_i}$, followed by $h_{C_{ij}}$ which encodes the $j$-th token from column $i$ since each column name may contain multiple tokens. Our context enhanced schema encoder (Figure~\ref{fig:contextreinforcing}) tries to learn a new representation $h_{C_i}$ for each column $i$ by strengthening the original encoder output with the global context information captured in $h_{\texttt{[CTX]}}$.

Denoting the number of tokens in column $i$ as $n_i$, the schema encoder summarizes each column by computing 
\begin{align}\label{eq:attnpooling}
h_{C_i} = \sum_{t=1}^{n_i} \alpha_{it} h_{C_{it}}
\end{align} 
where $\alpha_{it} \defeq \textsc{softmax}(s_{it})$. The alignment model $s_{it}$ tells how well $t$-th token of column $i$ matches the global context, and is defined as
\begin{align*}
    s_{it} &= f\Big(Uh_{\texttt{[CTX]}}/\sqrt{d}, Vh_{C_{it}}/\sqrt{d}\Big).
\end{align*}
Both $U, V \in \mathbf{R}^{m \times d}$, and we use simple dot product for $f$. 

While there is already some degree of context being captured in the output from sequence encoder, such influence is limited as self-attention tends to focus on only certain regions. On the other hand, the global contextual information captured in \texttt{[CTX]} is diverse enough, thus is used to complement the schema representation from sequence encoder.

Although context enhanced schema encoder and column attention introduced in~\citet{xu2017sqlnet} share a similar goal of better aligning natural language question and table schema, they differ significantly in both technical solution and the role played in the overall architecture. Column attention changes $h_{q_i}$ by signifying which query words are most relevant to each column. It does so for every column in the table, and columns are processed independent of each other. Context enhanced schema encoder, on the other hand, believes BERT style sequence encoder already performs well on the natural language side, and tries to come up with a better representation for schema. It uses only contextual information captured in \texttt{[CTX]} to update the schema part. Since \texttt{[CTX]} also contains information from other parts of the schema, columns are no longer updated independently.

\subsection{Output Layer}\label{sec:output}
The output layer composes the SQL program from both sequence encoder outputs $h_{\texttt{[CTX]}}, h_{q_1}, \cdots, h_{q_n}$ and context enhancing schema encoder outputs $h_{C_1}, h_{C_2}, \cdots, h_{\texttt{[EMPTY]}}$. Similar to \citet{xu2017sqlnet} and \citet{Hwang2019}, the task is decomposed into 6 sub-tasks, each predicting a part of the final SQL program. Unlike their models, X-SQL enjoys a much simplified structure due to context enhancement.

We first introduce a task dependent sub-network that modulates the schema representation $h_{C_i}$ using context $h_{\texttt{CTX}}$. Specifically,
\begin{align}\label{eq:colrep}
    r_{C_i} &= \text{LayerNorm}\big(U'h_{\texttt{[CTX]}} + V'h_{C_i}\big).
\end{align}
Different from Equation \ref{eq:attnpooling}, this computation is done separately for each sub-task, to better align the schema representation with the particular part of natural language question that each sub-task should focus on. 

The first task, \textsc{S-COL}, predicts the column for the \textsc{SELECT} clause. The probability of column $C_i$ being chosen for the \textsc{SELECT} statement is modeled as
\begin{align*}
    p^{\textsc{S-COL}}(C_i) & = \textsc{softmax}\big(W^{\textsc{S-COL}}r_{C_i}\big)
\end{align*}
with $W^{\textsc{S-COL}} \in \mathbf{R}^{1 \times d}$.  Note, \textsc{S-COL} depends on $r_{C_i}$ only, as opposed to both query and schema in previous work.

The second task, \textsc{S-AGG}, predicts the aggregator for the column selected by \textsc{SELECT}. To enhance the intuition that aggregator depends on the selected column type, e.g. \textsc{MIN} aggregator doesn't go with a string typed column, we explicitly add column type embedding to the model. The probability of the aggregator is computed as

\vspace{-.4cm}
\begin{small}
\begin{align}\label{eq:s-agg}
    p^{\textsc{S-AGG}}(A_j | C_i) & = \textsc{softmax}\bigg(W^{\textsc{S-AGG}}[j, \vcentcolon] \times \\
    &\textsc{LayerNorm} \left(U''h_{[\texttt{CTX}]} + V'' h_{C_i} + E^T_{C_i}\right)\bigg) \nonumber
\end{align}
\end{small}
where $W^{\textsc{S-AGG}} \in \mathbf{R}^{6 \times d}$ with 6 being the number of aggregators. Different from other sub-tasks we use $h_{C_i}$ here rather than $r_{C_i}$, as we incorporate column type in a similar way to Equation~\ref{eq:colrep}. Type embedding $E^T_{C_i}$ is learned separately from the one used by the sequence encoder.

\begin{table*}[h]
\renewcommand\thetable{2} 
\caption{Dev/test results for each sub-module. $\ast$ means results obtained by running SQLova code from github.}
\label{table:submodule}
\small\centering
\begin{tabular}{lccccccc}
\toprule
Model & \textsc{s-col} & \textsc{s-agg} & \textsc{w-num} & \textsc{w-col} & \textsc{w-op} & \textsc{w-val} \\
\midrule
SQLova & 97.3 / 96.8 & 90.5 / 90.6 & 98.7 / 98.5 & 94.7 / 94.3 & 97.5 / 97.3 & 95.9 / 95.4\\
X-SQL & 97.5 / 97.2 & 90.9 / 91.1 & 99.0 / 98.6	& 96.1 / 95.4 & 98.0 / 97.6 & 97.0 / 96.6\\
\midrule
SQLova + EG$^{\ast}$ & 97.3 / 96.5 & 90.7 / 90.4 & 97.7 / 97.0 & 96.0 / 95.5 & 96.4 / 95.8 & 96.6 / 95.9\\
X-SQL + EG & 97.5 / 97.2 & 90.9 / 91.1 & 99.0 / 98.6	& 97.7 / 97.2 & 98.0 / 97.5 & 98.4 / 97.9\\
\bottomrule
\end{tabular}
\end{table*}

The remaining 4 tasks \textsc{W-NUM}, \textsc{W-COL}, \textsc{W-OP} and \textsc{W-VAL} together determine the \textsc{WHERE} part. Task \textsc{W-NUM} finds the number of where clauses using $W^{\textsc{W-NUM}}h_{\texttt{[CTX]}}$, and is modeled as a classification over four possible labels each representing 1 to 4 where clauses in the final SQL. It doesn't predict the empty where clause case, which is delegated to \textsc{W-COL} through the Kullback-Leibler divergence as explained in Section~\ref{sec:training}. Task \textsc{W-COL} outputs a distribution over columns using 
\begin{align}\label{eq:wherecolumn}
    p^{\textsc{W-COL}}(C_i) & = \textsc{softmax}\big(W^{\textsc{W-COL}}r_{C_i}\big)
\end{align}
and based on the number from \textsc{W-NUM}, top scoring columns are selected for the where clauses. Task \textsc{W-OP} chooses the most likely operator for the given where column using $p^{\textsc{W-OP}}(O_j | C_i) = \textsc{softmax}\big(W^{\textsc{W-OP}}[j, \vcentcolon] r_{C_i}\big)$ . Intuitively operator also depends on the column type, and could be modeled in the same way as Task \textsc{S-AGG} in Equation~\ref{eq:s-agg}. However, we didn't observe improvement during experiments, therefore we prefer to keep the original simple model. Model parameters $W^{\textsc{W-NUM}}$, $W^{\textsc{W-COL}}$, $W^{\textsc{W-OP}}$ are in $\mathbf{R}^{4 \times d}$, $\mathbf{R}^{1 \times d}$ and $\mathbf{R}^{3 \times d}$ respectively, with number of possible operators being 3. 

Predicting value for where clause (task \text{W-VAL}) is formulated as predicting a span of text from query, which simply becomes predicting the beginning and the end position of the span using
\begin{align*}
    p^{\textsc{W-VAL}}_{\text{start}}(q_j|C_i) & = \footnotesize{\textsc{SOFTMAX}}\ \ g\big(U^{\text{start}}h_{q_j} + V^{\text{start}}r_{C_i}\big)
\end{align*}
and
\begin{align*}
    p^{\textsc{W-VAL}}_{\text{end}}(q_j|C_i) & = \footnotesize{\textsc{SOFTMAX}}\ \ g\big(U^{\text{end}}h_{q_j} + V^{\text{end}}r_{C_i}\big)
\end{align*}
where $g(x) \defeq Wx +b$. Parameters $U^{\text{start}}, V^{\text{start}}, U^{\text{end}}, V^{\text{end}} \in \mathbf{R}^{m \times d}$ and different $g$ functions are learned for predicting start and end.

\subsection{Training and Inference}\label{sec:training}
During training, we optimize the objective which is a summation over individual sub-task losses. We use cross entropy loss for task $\textsc{S-COL}$, $\textsc{S-AGG}$, $\text{W-NUM}$, $\textsc{W-OP}$ and $\textsc{W-VAL}$. The loss for $\textsc{W-COL}$ is defined as the Kullback-Leibler (KL) divergence between $D(Q || P^{\textsc{W-COL}})$, where $P^{\textsc{W-COL}}$ is modeled by Equation \ref{eq:wherecolumn}. Distribution $Q$ from ground truth is computed as follows:
\begin{itemize}
    \item If there is no where clause, $Q_{\texttt{[EMPTY]}}$ receives probability mass 1 for special column \texttt{[EMPTY]},
    \item For  $n\geq 1$ where clauses, each where column receives a probability mass of $\frac{1}{n}$.
\end{itemize}

Inference is relatively straightforward except for the $\textsc{W-COL}$. If the highest scoring column is the special column \texttt{[EMPTY]}, we ignore the output from \textsc{W-NUM} and return empty where clause. Otherwise, we choose top $n$ non-\texttt{[EMPTY]} columns as indicated by \textsc{W-NUM} and \textsc{W-COL}.

\section{Experiments}

We use the default train/dev/test split of the WikiSQL dataset. Both logical form accuracy (exact match of SQL queries) and execution accuracy (ratio of predicted SQL queries that lead to correct answer) are reported. The logical form accuracy is the metric we optimize during training.

Table \ref{table:main} includes results both with and without execution guidance (EG) applied during inference~\citep{Wang2018}. We compare our results with the most recent work of WikiSQL leaderboard, including the previous state-of-the-art SQLova model. X-SQL is shown to be consistently and significantly better across all metrics and achieves the new state-of-the-art on both dev and test set. 
Without EG, X-SQL delivers an absolute 2.6\% (83.3 vs. 80.7) improvement in logical form accuracy and 2.5\% improvement in execution accuracy on test set. Even with EG, X-SQL is still 2.4\% better in logical form accuracy, and 2.2\%  better in execution accuracy. It is worth noting that X-SQL+EG is the first model that surpasses the 90\% accuracy on test set. On the other hand, for dev set human performance is estimated to be 88.2\% according to \citet{Hwang2019}. X-SQL is the first model better than human performance without the help of execution guidance. 

\begin{table}[h]
\renewcommand\thetable{1} 
\caption{Logical form (\textit{lf}) and execution (\textit{ex}) accuracy.}
\label{table:main}
\small\centering
\begin{tabular}{lcccc}
\toprule
\multirow{2}{*}{Model} & \multicolumn{2}{c}{Dev} & \multicolumn{2}{c}{Test} \\
\cmidrule{2-5}
&  Acc\textsubscript{lf} &  Acc\textsubscript{ex} & Acc\textsubscript{lf} & Acc\textsubscript{ex}\\
\midrule
SQLNet & 63.2 & 69.8 & 61.3 & 68.0 \\
SQLova & 81.6 & 87.2 & 80.7 & 86.2 \\
X-SQL &		\textbf{83.8} &  \textbf{89.5} & \textbf{83.3} & \textbf{88.7}\\
\midrule
SQLova + EG & 84.2 & 90.2 & 83.6 & 89.6 \\
X-SQL +  EG & \textbf{86.2} & \textbf{92.3} & \textbf{86.0} & \textbf{91.8}\\
\bottomrule
\end{tabular}
\end{table}

Table~\ref{table:submodule} reports the accuracy for each sub-task, and demonstrates consistent improvement. 
In particular, task \textsc{W-COL} shows an absolute 1.1\% gain without EG and 1.7\% with EG. We attribute this to our new approach of formalizing the where column prediction as a list-wise ranking problem using KL divergence. 
Another significant improvement is the \textsc{W-VAL} task, with an absolute 1.2\% gain without EG and 2.0\% with EG. This partly results from the column set prediction (i.e. \textsc{W-COL}) improvement as well, since the value generation  depends highly on the predicted column set for the where clause.

\section{Conclusion}
We propose a new model X-SQL, demonstrate its exceptional performance on the WikiSQL task, and achieve new state-of-the-art across all metrics. 
While the contribution around loss objective may be bounded by the specific SQL syntax that WikiSQL uses, how contextual information is leveraged and how schema type is used can be immediately applied to other tasks that involve pre-trained language model for structured data. Future work includes experimenting with more complex dataset such as Spider~\cite{Yu2018EMNLP}.


\newpage

\bibliography{emnlp-ijcnlp-2019}
\bibliographystyle{acl_natbib}

\end{document}